\documentclass[journal]{IEEEtran}
\usepackage{cite}

\ifCLASSINFOpdf
   \usepackage[pdftex]{graphicx}
\else
\fi
\usepackage{url}

\usepackage{hyperref}
\usepackage{amsmath,amsfonts}
\usepackage{booktabs}       
\usepackage{multirow, makecell}
\usepackage{footmisc}
\usepackage{pifont}
\usepackage{textcomp}
\usepackage{xcolor}

\DeclareMathSymbol{\shortminus}{\mathbin}{AMSa}{"39}

\newcommand{\cmark}{\ding{51}}%
\newcommand{\xmark}{\ding{55}}%

\newcommand{\Sref}[1]{\S\ref{#1}}

\newcommand{\Fref}[1]{Fig.~\ref{#1}}

\newcommand{\Tref}[1]{Table~\ref{#1}}

\begin{document}
%
\title{LegoNN: Building Modular Encoder-Decoder Models}
%
%
%

\author{Siddharth~Dalmia, 
        Dmytro~Okhonko, 
        Mike~Lewis, 
        Sergey~Edunov, 
        Shinji~Watanabe,~\IEEEmembership{Fellow,~IEEE,}
        Florian~Metze,~\IEEEmembership{Fellow,~IEEE,}
        Luke~Zettlemoyer, 
        and~Abdelrahman~Mohamed,~\IEEEmembership{Member,~IEEE}
\thanks{This work was done at FAIR, Meta AI.} 
\thanks{Siddharth Dalmia and Shinji Watanabe are affiliated with LTI at Carnegie Mellon University. Florian Metze is affiliated with both Language Technologies Institute at Carnegie Mellon University and Meta AI. Abdelrahman Mohamed is affiliated with Rembrand Inc. Dmytro Okhonko is affiliated with Samaya AI. All other authors are affiliated with Meta AI.}
}
\maketitle

\begin{abstract}
State-of-the-art encoder-decoder models (e.g.~for machine translation (MT) or automatic speech recognition (ASR)) are constructed and trained end-to-end as an atomic unit. No component of the model can be (re-)used without the others, making it impossible to share parts, e.g.~a high resourced decoder, across tasks. We describe LegoNN, a procedure for building encoder-decoder architectures in a way so that its parts can be applied to other tasks without the need for any fine-tuning. To achieve this reusability, the interface between encoder and decoder modules is grounded to a sequence of marginal distributions over a pre-defined discrete vocabulary.
We present two approaches for ingesting these marginals; one is differentiable, allowing the flow of gradients across the entire network, and the other is gradient-isolating. 
To enable the portability of decoder modules between MT tasks for different source languages and across other tasks like ASR, we introduce a modality agnostic encoder which consists of a length control mechanism to dynamically adapt encoders' output lengths in order to match the expected input length range of pre-trained decoders.
We present several experiments to demonstrate the effectiveness of LegoNN models: a trained language generation LegoNN decoder module from German-English (De-En) MT task can be reused without any fine-tuning for the Europarl English ASR and the Romanian-English (Ro-En) MT tasks, matching or beating the performance of baseline. After fine-tuning, LegoNN models improve the Ro-En MT task by 1.5 BLEU points and achieve 12.5\% relative WER reduction on the Europarl ASR task. 
To show how the approach generalizes, we compose a LegoNN ASR model from three modules -- each has been learned within different end-to-end trained models on three different datasets -- achieving an overall WER reduction of 19.5\%.
\end{abstract}

\begin{IEEEkeywords}
end-to-end, encoder-decoder models, modularity, speech recognition, machine translation
\end{IEEEkeywords}

%
\IEEEpeerreviewmaketitle

\section{Introduction}
\label{sec:intro}

\begin{figure}[t]
  \centering
    \includegraphics[width=\linewidth]{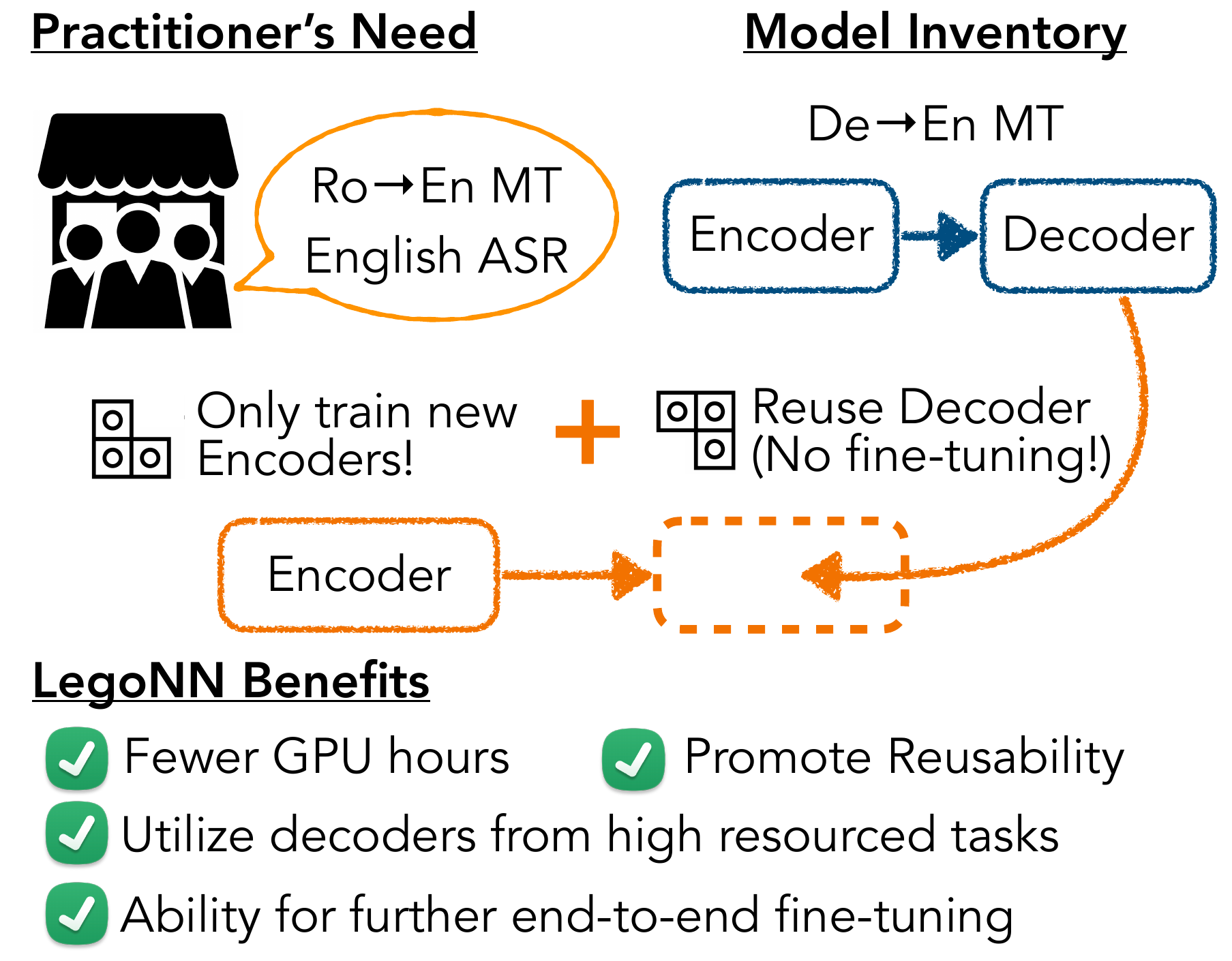}
    \caption{\textbf{LegoNN Framework}: Building encoder-decoder models in the LegoNN framework, allows practitioners to reuse components like decoder modules for various sequence prediction tasks. For example, in this figure, an English predicting decoder from a German-English machine translation system can be re-used for both Romanian-English machine translation and English speech recognition without any fine-tuning steps. This saves overall compute resources, promotes re-usability, and allows practitioners to utilize decoders from high-resourced tasks for under-resourced ones. The composed model is end-to-end differentiable leaving room for further improvements through fine-tuning.}
    \label{fig:pg1}
\end{figure}

\IEEEPARstart{T}{raining} end-to-end models for machine translation (MT) or automatic speech recognition (ASR) require learning of multiple implicit functions~\cite{bahdanau2014neural, sutskever2014sequence, vaswani2017attention,chan2016listen,bahdanau2016end}. An MT model is implicitly doing both word translation and language generation, while an ASR model combines phoneme recognition, pronunciation modeling, and language generation. 
These fully differentiable models are conceptually simple and work well in practice. However, they forgo opportunities to share common logical functions between different tasks, such as their decoders.
Training such \emph{monolithic} models leads to wasted compute during training and less interpretable architectures overall due to the lack of re-usability of trained components that perform the same logical function across tasks. 

Motivated by modularity principles in software design ~\cite{Baldwin_99} where modules have interpretable interfaces and are reusable within other programs, we seek to build an AI paradigm where \emph{trained} components (modules) of an end-to-end model can be re-used across different tasks (programs) provided they perform the same function (interface).
With a focus on sequence prediction tasks, we introduce LegoNN, a procedure for constructing encoder-decoder models where trained components such as decoder modules can be reused across various sequence tasks such as MT, ASR, or Optical Character Recognition (OCR). 
As summarized in \Fref{fig:pg1}, for AI models, enforcing modularity helps save computing resources by reusing components and helps build systems for under-resourced tasks by utilizing shareable components from higher resourced tasks. Additionally, having interpretable interfaces enable monitoring the performance of individual encoder or decoder modules and their contributions to the overall end-to-end performance.

More concretely, in our LegoNN encoder-decoder framework, encoders have an interpretable interface by outputting a sequence of distributions over a discrete vocabulary, 
derived from the final output labels (e.g. phonemes or sub-words). During training, we add an additional Connectionist Temporal Classification (CTC) loss~\cite{graves2006connectionist} on the encoder outputs to enforce this modularity (\Sref{sec:vocab}). Our decoder modules are extended with Ingestor layers (ING) that accept these distributions as inputs (\Sref{sec:ing}). We experiment with two types of ingestor layers: a differentiable Weighted Embedding (WEmb) ingestor allowing gradient flow across the entire network, and a gradient-isolating Beam Convolution (BeamConv) one. Given that LegoNN decoder modules can be trained for one MT task and then reused for another MT task with a different source language or for a sequence task with the same target language such as ASR or OCR, we propose a modality agnostic encoder for sequence prediction task that uses an output length controller (OLC) unit to adapt any input modality to a sequence of encoder representations that matches the expected input length of another (\Sref{sec:olc}). LegoNNs, as also demonstrated in our experiments, enable sharing of trained decoders \footnote{Our work on LegoNNs is orthogonal to the recent work on sharing pre-trained encoders, e.g BERT \cite{devlin-bert}. Combining the benefits of these two complementary approaches is left for future work.} and intermediate modules between different tasks and domains without jointly training for both tasks and no fine-tuning steps. The composed LegoNN model preserves end-to-end differentiability of each individual system, allowing room for further improvements through fine-tuning.

Our experiments show that we can achieve modularity without sacrificing performance. On the standard large scale En-De WMT and Switchboard ASR benchmarks, LegoNN models reach levels of performance competitive with the standard monolithic architectures (\Sref{sec:exp_performance}), while still passing our stress tests for testing modularity (\Sref{sec:resilience}). 
We show the value of modularity by seamlessly composing a decoder module trained within a German-English (De-En) WMT system with other pre-trained encoder modules from different MT and ASR systems, without any joint training or fine-tuning, to match or beat generation performance for the Europarl English ASR task and the Romanian-English (Ro-En) WMT task (\Sref{sec:exp_crosstasks}). When such composed LegoNN models are fine-tuned for a few thousand steps towards the target domain, they improve the Ro-En MT task by 1.5 BLEU points and improve on the Europarl ASR task by 12.5\% WER relative to the baseline system (\Sref{sec:fine_tune}).
To demonstrate the flexibility of reusing LegoNN modules, we construct an ASR system that is composed of modules that have been trained independently on three different tasks, without performing any fine-tuning and with almost no performance degradation. The modules are: (1) a phoneme recognizer from the Europarl ASR model, (2) a pronunciation model from the TED-LIUM ASR model, and (3) a language generation decoder from the WMT model (\Sref{sec:exp_crosstasks}). With a few end-to-end fine-tuning steps, the composed model beats the baseline Europarl ASR model by 19.5\% relative WER, also improving over the previously composed LegoNN model for this task (\Sref{sec:fine_tune}).

\section{Background}
\label{sec:current}
\subsection{Cross-Entropy Loss in Encoder-Decoder Models} Given an input sequence of length T, $X_{1:T}$ and an output sequence, $Y$, which can be factorized into a sequence of tokens, $L$, of length N, $L_{1:N} = \{L_n \in \mathcal{V}_\text{Attn} | n\in\{1:N\},L_{1:N}=Y\}$. The encoder-decoder model \cite{oxford_s2s_2013, sutskever2014sequence, bahdanau2014neural} models the likelihood ($\mathbb{P}^{\mathcal{V}_\text{Attn}}$) for the prediction of the next token ($L_n$) given the previous tokens ($L_{1:n\shortminus1}$) and the input sequence ($X_{1:T})$, 
\begin{align}
\mathbb{P}^{\mathcal{V}_\text{Attn}}_n = P( \cdot \mid X_{1:T}, L_{1:n\shortminus1}), \label{regular_decoder}
\end{align}
where $\mathbb{P}_n^{\mathcal{V}_\text{Attn}}$ is the posterior distribution for predicting the next token $L_n$ at position $n$ defined over the vocabulary space $\mathcal{V}_\text{Attn}$ from $n = \{1:N\}$.
They are trained by minimizing the token-level cross-entropy ($\mathcal{F}_{\text{CE}}$) loss between the true tokens $L_{1:N}$ and the decoder predicted distributions ($\mathbb{P}^{\mathcal{V}_{\text{Attn}}}_{1:N}$),
\begin{align}
\mathbf{h}^E_{1:T} &= \operatorname{encoder}(X_{1:T}), \label{subsampling_ignore}\\
\mathbb{P}_{n}^{\mathcal{V}_{\text{Attn}}} &= \operatorname{softmax}(\operatorname{decoder}(\mathbf{h}^E_{1:T}, L_{1:n\shortminus1})), \label{dec_eq1}\\
\mathcal{F}_{\text{CE}}(L_{1:N},\mathbb{P}_{1:N}^{\mathcal{V}_{\text{Attn}}}) &= -\log \left(\prod_{n=1}^{N} \mathbb{P}_{n}^{\mathcal{V}_{\text{Attn}}}(y=L_n) \right), \label{ce_eq}
\end{align}
where $\mathbb{P}_{n}^{\mathcal{V}_{\text{Attn}}}(y=L_n)$ is the probability of predicting the ground truth token $L_n$ at n-th decoder step.
To avoid an explosion in notations and maintain consistency between speech and text based encoder-decoder models, we avoid denoting any changes to sequence lengths in Eq. (\ref{subsampling_ignore}) due to sub-sampling in speech encoders \cite{chan2016listen}. 

\subsection{Connectionist Temporal Classification Loss} 
For the input sequence $X_{1:T}$ and the output sequence $Y$, which is factorized into a sequence of tokens $L$ of length $N$, $L_{1:N} = \{ L_n \in \mathcal{V}_\text{CTC} | n \in \{1:N\}, L_{1:N} = Y\}$. CTC model \cite{graves2006connectionist} models the likelihood ($\mathbb{P}^{\mathcal{V}_{\text{CTC}}}$) of producing a valid $X \rightarrow L$ alignment for each input step $t$,
\begin{align}
   \mathbb{P}^{\mathcal{V}_{\text{CTC}}}_{t} = P( \cdot \mid X_{1:T}), \label{eq:ctc_likelihoods}
\end{align}
where $\mathbb{P}^{\mathcal{V}_{\text{CTC}}}_{t}$ is the conditionally independent posterior distribution for predicting the $X \rightarrow L$ alignment at time step $t$ defined over the vocabulary space $\mathcal{V}_\text{CTC}$ from $t = \{1:T\}$. They are trained using CTC loss \cite{graves2006connectionist} that minimizes the negative conditional likelihood of all possible monotonic alignments, $Z \in \mathcal{Z}(T,L)$, between sequence $X$ and $L$ where each $Z = \{ z_t \in l | t \in \{1:T\}, l \in \mathcal{V}_\text{CTC} \}$ such that $Z$ is a valid alignment for producing $L$,
\begin{align}
    \mathbb{P}^{\mathcal{V}_{\text{CTC}}}_{1:T} &=  \operatorname{softmax}\left( \operatorname{encoder}(X_{1:T})*W_o \right), \label{eq:ctc_out}\\
    \mathcal{F}_{\text{CTC}}(L_{1:N},\mathbb{P}^{\mathcal{V}_{\text{CTC}}}_{1:T}) &= -\log \sum_{Z \in \mathcal{Z}(T,L)} \left(\prod_{t=1}^{T} \mathbb{P}^{\mathcal{V}_{\text{CTC}}}_t (z = z_t ) \right), \label{eq:ctc_loss_fn}
\end{align}
where $\mathbb{P}_t^{\mathcal{V}_\text{CTC}}( z = z_t) $ is the probability of predicting the output token $z_t \in \mathcal{V}_\text{CTC}$ for the t-th input step. $W_o \in \mathbb{R}^{d \times |\mathcal{V}_{\text{CTC}}|}$ projects the encoder representations into the output vocabulary space of $\mathcal{V}_{\text{CTC}}$ and $\mathbb{P}^{\mathcal{V}_{\text{CTC}}}_{1:T} \in \mathbb{R}^{T \times |\mathcal{V}_{\text{CTC}}|}$ is the locally-normalized per input step probabilities of the output alignment. 
We omit the extra CTC blank symbol in the equation above for clarity of presentation; see \cite{graves2006connectionist} for more details. The marginalization sum is efficiently computed using dynamic programming. 

\section{LegoNN for Modular Encoder-Decoder Models}
\label{sec:proposed}

We propose decomposing encoder-decoder models into one (or more) encoder modules followed by an auto-regressive decoder module, which can be reused for other tasks and domains, without sacrificing its end-to-end differentiability. Each module produces a sequence of marginal distributions over a pre-defined discrete vocabulary, which is consumed by an ingestor component in the subsequent module. Each encoder module can be made modality agnostic by using an output length controller unit that matches the input sequence length to the expected length of the subsequent module, allowing re-usability between similar tasks from different modalities. LegoNN modules trained either jointly or independently can be reused for new tasks without the need for any fine-tuning. 

The LegoNN procedure introduces three operations that we discuss in detail in this section, (1) designing an interpretable interface between modules (\Sref{sec:vocab}), (2) output length controller unit in the LegoNN encoder (\Sref{sec:olc}) and (3) ingestor of probability distributions in the LegoNN decoder (\Sref{sec:ing}). We also provide some additional insights into training LegoNN systems (\Sref{sec:insights_training}).

\subsection{Designer-defined module interface}
\label{sec:vocab}

We look at encoder-decoder models as full software programs executing the specific function of mapping one sequence of input symbols or vectors to another. While some components of software programs, i.e. libraries, may be reused for numerous future programs, {\em trained} components of the traditional encoder-decoder models \cite{sutskever2014sequence, bahdanau2014neural}, like decoders, are not designed to be reused independently with other models or tasks.~\footnote{While deep learning toolkits have a modular code design where the code for similar model architectures can be re-used to train new components for different tasks \cite{tensorflow2015-whitepaper,jax2018github,paszke2019pytorch}. In this work, we aim to make the entire component of a {\em trained} model re-usable across tasks, which includes re-using the code, parameters, and input/output space of the components.}  
Well-defined and abstract input/output interfaces are prerequisites for developing reusable software libraries, however, encoder-decoder models fall short in this respect.

In order to have modular software properties between encoder and decoder modules, we propose that the interface between modules can be defined as distributions over a fixed categorical space while capturing similar information as the encoder from a standard monolithic encoder-decoder model. We define our interface as a sequence of conditionally independent distributions over a vocabulary space learned using the CTC loss for each input step (\Sref{sec:interface_enc_dec}). We then extend this approach to introduce more than one modularity point in an encoder-decoder model by having a sequence of encoder modules, each with defined functionalities and interface (\Sref{sec:interface_multiple_enc}). In order for the distribution over the vocabulary interface to be effective, we also discuss the importance of generalizability of the interface for future tasks (\Sref{sec:vocab_generalizability}).

\begin{figure*}[t]
  \centering
    \includegraphics[width=\linewidth]{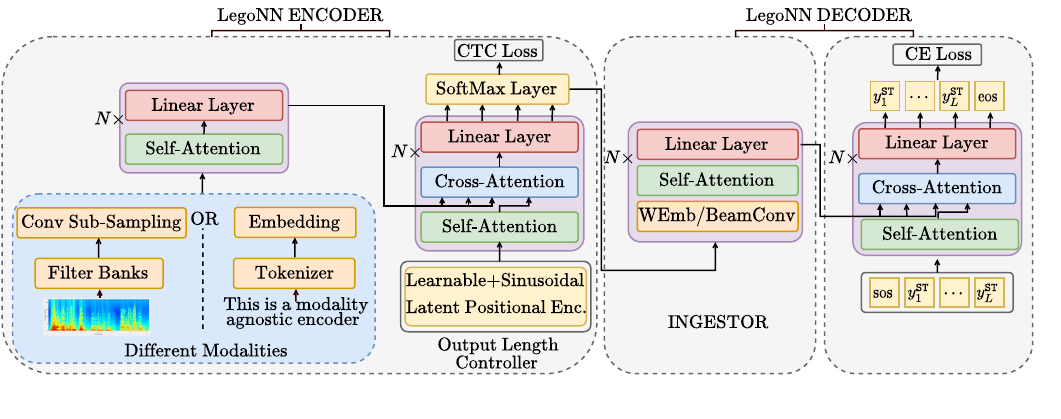}
    \caption{\textbf{LegoNN Encoder-Decoder Model}: This figure presents the schematics and the information flow in our LegoNN encoder-decoder model. LegoNN decomposes encoder-decoder models into reusable modules by (1) grounding module output into a distribution over an interpretable vocabulary using softmax output in LegoNN encoder guided using CTC Loss; (2) adding an ingestor (ING) to process input marginal distributions in the LegoNN decoder; and (3) an output length controller (OLC) in the LegoNN encoder to match the input length of subsequent modules.}
    \label{fig:lego_model}
\end{figure*}

\subsubsection{Interpretable interfaces between encoders and decoders}
\label{sec:interface_enc_dec}
Encoders communicate with decoders through a sequence of continuous hidden representations (Eq. \ref{dec_eq1}) that are by-products of the end-to-end model optimization process. 
An individual unit in these hidden vectors does not have physical meaning. Their properties can vary across random seeds and training hyper-parameters even when the same task, model, and training data are used. As we see from Eq. (\ref{regular_decoder}) and (\ref{dec_eq1}), although the decoder is conditioned on encoder states ($\mathbf{h}_{1:T}^{E}$) the encoder-decoder still directly models the conditional next token prediction over the input as these encoder states tightly coupled into the model and have no physical meaning. 

To enable reusability between different tasks and models, LegoNN grounds encoder outputs into a distribution over a discrete vocabulary space that is pre-defined by the model designer. For applications such as ASR and MT, such intermediate discrete vocabulary can be defined over phonemes, or some byte-pair encoding (BPE) dictionary \cite{BPE, sentencepiece} driven from the task labels, which does not need to be the same as the model's output dictionary. Such an encoder can be designed similarly to Eq. (\ref{eq:ctc_out}), to produce locally-normalized output distributions for each encoder position ($\mathbb{P}^{\mathcal{V}_{\text{CTC}}}_{1:T}$) learned using CTC loss (Eq. \ref{eq:ctc_loss_fn}), where $\mathcal{V}_\text{CTC}$ is the intermediate discrete vocabulary of the interface. 
Following the notations introduced in \Sref{sec:current}, a LegoNN encoder-decoder model effectively models the likelihood for the next token prediction given the previous tokens ($L_{1:n\shortminus1}$), and the encoder states ($\mathbb{P}^{\mathcal{V}_{\text{CTC}}}_{1:T}$),
\begin{equation}
\mathbb{P}^{\mathcal{V}_\text{Attn}}_n = P( \cdot \mid \mathbb{P}^{\mathcal{V}_{\text{CTC}}}_{1:T}, L_{1:n\shortminus1}). \label{lego_decoder}
\end{equation}
Note how the conditional is different from Eq. (\ref{regular_decoder}). Here, the encoder states ($\mathbb{P}^{\mathcal{V}_{\text{CTC}}}_{1:T}$) are not just hidden transformations of the input ($X_{1:T}$), but distributions over an intermediate vocabulary of the interface, $\mathcal{V}_\text{CTC}$ defined over the output sequence $Y$ and are modeled by the CTC loss given input,
\begin{equation}
\mathbb{P}^{\mathcal{V}_{\text{CTC}}}_{1:T} = P( \cdot \mid X_{1:T}). \label{lego_encoder_ctc}
\end{equation}
In order to model the two distributions in Eq. (\ref{lego_decoder}) and (\ref{lego_encoder_ctc}), we can train LegoNN models with a supervised CTC loss applied over the encoder output distributions (Eq. \ref{eq:ctc_loss_fn}) along with the decoder token-level cross-entropy loss (Eq. \ref{ce_eq}),
\begin{equation}
\mathcal{F}_{\text{obj}} = \mathcal{F}_{\text{CE}}( L^{\mathcal{V}_\text{Attn}}, \mathbb{P}^{\mathcal{V}_{\text{Attn}}}) + \mathcal{F}_{\text{CTC}}(L^{\mathcal{V}_\text{CTC}}, \mathbb{P}^{\mathcal{V}_{\text{CTC}}}), \label{eq_single_lego_module}
\end{equation}
where $L^{\mathcal{V}_\text{Attn}}$ and $L^{\mathcal{V}_\text{CTC}}$ are target sequences tokenized using the vocabularies $\mathcal{V}_\text{Attn}$ and $\mathcal{V}_\text{CTC}$.
This modeling framework allows the decoder module to accept any encoder that produces the same distribution as the decoder was trained towards ($\mathbb{P}^{\mathcal{V}_{\text{CTC}}}_{1:T}$), thereby building an interpretable interface to enforce modularity between the encoder-decoder modules. Such modeling also gives the encoder and decoder modules in the LegoNN framework specific functionalities toward the target task. For example, an MT encoder module is not expected to solve the overall translation task but rather acts as a word/phrase translation component or an ASR encoder acts as a phoneme/sub-word recognizer whose outputs are refined using an auto-regressive decoder.

While the softmax operation in Eq. (\ref{eq:ctc_out}) produces an intermediate distribution over the defined vocabulary and the model can be trained with only the decoder cross-entropy loss, it is still essential to enforce the desired encoder distributions using CTC loss. Experiments in \Sref{sec:exp_encoder_grounding} show that without CTC loss at the encoder and only training with the decoder cross-entropy loss, the encoder-decoder models are not modular. Due to the marginalization of all output alignments (Eq. \ref{eq:ctc_loss_fn}), CTC loss produces conditionally independent output distributions for each input step that is dependent only on the input (Eq. \ref{eq:ctc_likelihoods}). This ensures that the distributions produced by the CTC loss have no output label bias \cite{awnilabelbias} and represent a transformation function of only the input, making it possible to capture similar information as that of an encoder from a monolithic encoder-decoder model. \footnote{Local normalization using the softmax operation at each input step can suppress information such as confidence of a state, whereas hidden representations in monolithic encoders preserve it~\cite{bottou1997global}.}

\subsubsection{Introducing multiple modularity points}
\label{sec:interface_multiple_enc}

A LegoNN model is not restricted to containing only two modules; an encoder and a decoder. There may be a sequence of encoder modules, each designed to perform a certain function through their respective pre-defined output vocabulary. The ASR system is one example whose encoder can be divided into two modules, a phoneme recognizer, and a pronunciation model, followed by the auto-regressive language generating module. In this case, the CTC loss is applied more than once, and training of LegoNN models can be extended from Eq. (\ref{eq_single_lego_module}) to a joint loss over each module,
\begin{equation}
\mathcal{F}_{\text{obj}} = \mathcal{F}_{\text{CE}}( L^{\mathcal{V}^{(M)}_\text{Attn}}, \mathbb{P}^{\mathcal{V}^{(M)}_{\text{Attn}}}) + \sum_{i=1}^{M-1} \mathcal{F}_{\text{CTC}}(L^{\mathcal{V}^{(i)}_\text{CTC}}, \mathbb{P}^{\mathcal{V}^{(i)}_{\text{CTC}}}), \label{eq_multi_module}
\end{equation}
where $M$ is the total number of modules in the LegoNN system, $\mathcal{V}^{i}$ is the pre-defined vocabulary at the interface of module $i$. $\mathbb{P}^{\mathcal{V}^{(i)}}$ is the predicted distribution over vocabulary $\mathcal{V}^{i}$ at module $i$ and $L^{\mathcal{V}^{(i)}}$ is the target sequence tokenized using vocabulary $\mathcal{V}^{(i)}$. All modules in the LegoNN system have a CTC loss at their output, except the final auto-regressive decoder module with a token-level CE loss.

Furthermore, applying a supervised loss at the output of each module brings two extra benefits: (a) It extends our ability to evaluate and diagnose the performance on multiple points across the model. This can guide modeling decisions, e.g. adding more modeling capacity to one module over the other. and (b) An encoder module may be trained in conjunction with its decoder or independently by itself. For an ASR task, this may be useful when getting access to more audio training data or adapting the system to new acoustic conditions where retraining the decoder would be wasteful or even hurtful towards generalization. We present these properties through experiments in \Sref{appendix_inferface} and \Sref{sec:lowresource_legoNN}. 

\subsubsection{Defining vocabularies that generalize across tasks}
\label{sec:vocab_generalizability}
As with software libraries, designing a module interface with the right level of generality is challenging but allows for wider reuse of modules across tasks and domains. 
To reuse a decoder module from an MT LegoNN model to an ASR task, the output vocabulary of the speech encoder must be compatible with the input vocabulary of the translation decoder.
This is true even within a single task -- for example in ASR, a vocabulary of phonemes developed for a phoneme recognizer module in a read speech dataset, e.g. audiobooks, may not be the best one for spontaneous conversational situations which are full of hesitations and false starts. We propose designing a shared vocabulary by combining target units from multiple potential future tasks and finding a vocabulary at their intersection. 

\subsection{LegoNN Encoder-Decoder model}
\Fref{fig:lego_model} provides the schematics of our proposed LegoNN encoder-decoder model, which follows the modeling framework described in the section above. The model is designed to work for various sequence prediction tasks with various input modalities such as speech or text. The designed LegoNN modules take into account reusing LegoNN decoder modules across different tasks such as MT and ASR. For this purpose, we designed a modality agnostic encoder with an output length controller unit (\Sref{sec:olc}), which we call the LegoNN encoder as shown in the left side of \Fref{fig:lego_model}. The LegoNN decoder, as shown on the right side of \Fref{fig:lego_model}, is modified with an added ingestor component (\Sref{sec:ing}) which would consume the distributions produced by the LegoNN encoder. We have detailed the individual modules in the following sections.

\subsubsection{LegoNN Encoder}
\label{sec:olc}
For an input $X_{1:T}$, which can either be filter banks for speech frames or embeddings for text tokens, the LegoNN encoder consists of two sets of repeating blocks, which we call (1) the modality encoder (\Sref{para:modality_encoder}) and (2) the output length controller unit (\Sref{para:olc}). 

\paragraph{Modality Encoder}
\label{para:modality_encoder}
The modality encoder is just like a regular transformer encoder \cite{vaswani2017attention} that simply encodes the input context through repeating blocks of multi-head self-attention, $\operatorname{MHA}(x,x,x)$, and position-wise feedforward layers, $\operatorname{FFN}(x)$ \cite{vaswani2017attention}. For each $X \in X_{1:T}$, 

\begin{align}
\tilde{X} &= \operatorname{LayerNorm}(X), \\
X &=X+\operatorname{MHA}(\tilde{X},\tilde{X}_{1:T},\tilde{X}_{1:T}), \label{eq_enc1} \\
X &=X+\operatorname{FFN}(\operatorname{LayerNorm}(X)). \label{eq_enc2}
\end{align}
These blocks are followed by a final $\operatorname{LayerNorm}(X)$ which returns an input context-aware representation of length $T$. In principle, the modality encoder learns a sequence of representations from any input modality and can also contain modality-specific architectures such as conformers for speech \cite{gulati2020conformer}, vision-transformers for images \cite{dosovitskiy2020image} and pre-trained language models for text \cite{devlin-bert}. As the lengths of the learned representations can be quite different for different modalities such as speech and text, we introduce the output length controller unit.

\paragraph{Ouput Length Controller (OLC) unit}
\label{para:olc}
One of the challenges of using modules across sequential tasks, where inputs and outputs have different lengths, is adapting the output length of an encoder module trained on one task to match the expected input length of another. For example, encoder modules from an ASR task encode inputs in more time steps compared to an MT encoder. Naive up- or down-sampling approaches, e.g., pooling or replicating time-steps \cite{libovicky-helcl-2018-end}, cover only integer length ratios that are either aggressively down-sampling or unnecessarily up-sampling output sequence lengths. To solve this problem, we introduce an Output Length Controller (OLC) component in LegoNN encoders to enable working with fractional length ratios between inputs and outputs of the same module. 

OLC is a novel application of cross-attention \cite{bahdanau2014neural, vaswani2017attention} between two groups of transformer layers in a multi-layer module. If the output of our modality encoder processes the input $X_{1:T}$ to produce input representations of length $T$ and we want to convert it into $K$ length. The OLC first initializes a sequence of $K$ positional embeddings, $\mathbf{h}^{\text{PE}}_{1:K}$,
\begin{align}
\mathbf{h}^{\text{PE}}_{1:K} &=\operatorname{SinusoidalPE}(1:K) + \operatorname{LearnablePE}(1:K),
\end{align}
where $\operatorname{LearnablePE}$ and $\operatorname{SinusoidalPE}$ are learnable and sinusoidal positional embeddings \cite{conv_seq2seq, vaswani2017attention}. 

The $\mathbf{h}^{\text{PE}}_{1:K}$ representations pass through another set of transformer blocks that applies an additional cross-attention operation \cite{bahdanau2014neural, vaswani2017attention} over the modality encoder representations. The blocks consists of multi-head self-attention, $\operatorname{MHA}(y,y,y)$, multi-head cross-attention, $\operatorname{MHA}(y,x,x)$, and position-wise feedforward, $\operatorname{FFN}(y)$. For each $\mathbf{h}^{\text{PE}} \in \mathbf{h}_{1:K}^{\text{PE}}$, 
\begin{align}
\tilde{\mathbf{h}}^{\text{PE}} &= \operatorname{LayerNorm}(\mathbf{h}^{\text{PE}}), \\
\mathbf{h}^{\text{PE}} &=\mathbf{h}^{\text{PE}}+\operatorname{MHA}(\tilde{\mathbf{h}}^{\text{PE}}, \tilde{\mathbf{h}}^{\text{PE}}_{1:K}, \tilde{\mathbf{h}}^{\text{PE}}_{1:K}), \\
\mathbf{h}^{\text{PE}} &=\mathbf{h}^{\text{PE}}+\operatorname{MHA}(\operatorname{LayerNorm}(\mathbf{h}^{\text{PE}}), X_{1:T}, X_{1:T}), \\
\mathbf{h}^{\text{PE}} &=\mathbf{h}^{\text{PE}}+\operatorname{FFN}(\operatorname{LayerNorm}(\mathbf{h}^{\text{PE}})).
\end{align}
These blocks are followed by a final $\operatorname{LayerNorm}(\mathbf{h}^{\text{PE}})$, which now returns a sequence of $K$ representations, $\mathbf{h}^{\text{PE}}_{1:K}$, that encodes the input, $X_{1:T}$. The length $K$ can be either up-sampled or down-sampled depending on the input modality. The sampling factor can be set to anything such that it returns at least a singular ($K=1$) unit sequence. In this paper, we set the sampling rate as a fixed factor to the input length that tries to match MT and ASR encoder output length by computing the most accommodating ratio using the input and target sequence lengths of samples from the training data such that the ratio also does not violate the CTC criterion \cite{graves2006connectionist} (\Sref{sec:exp_setup}).

\paragraph{CTC Interface}
Finally the encoder representations, $\mathbf{h}^{\text{PE}}_{1:K}$ is transformed into the encoder vocabulary size followed by a $\operatorname{softmax}$ operation, as shown in Eq. (\ref{eq:ctc_out}), to produce a distribution ($\mathbb{P}^{\mathcal{V}_{\text{CTC}}}_{1:K}$) over the encoder vocabularies. These distributions are passed to the LegoNN decoder and to the CTC loss computation. \\

\subsubsection{LegoNN Decoder with Ingestor of probability distributions}
\label{sec:ing}

\begin{figure*}[t]
  \centering
    \includegraphics[width=\linewidth]{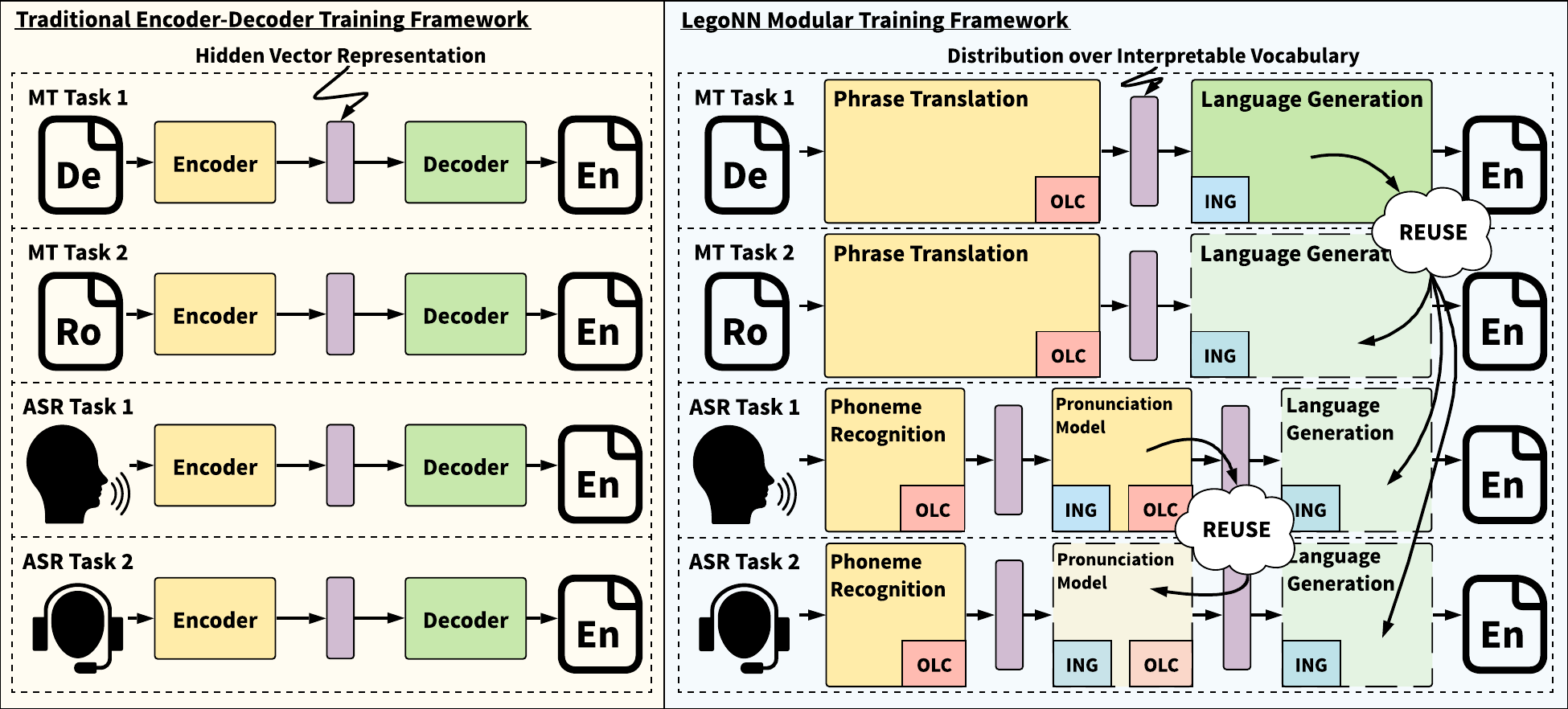}
    \caption{\textbf{Benefits of LegoNN}: Given a scenario where practitioners have a De-En MT system and want to build additional ASR and MT systems, with the LegoNN framework, they only need to build new encoder systems and can directly re-use decoder modules from their inventory. For example, re-using the De-En MT decoder for the Ro-En MT task and English ASR task. Additionally, when building an ASR system on a different domain they can re-use components from both ASR and MT systems like the pronunciation module from the previous ASR system and the decoder module from the De-En MT system.}
    \label{fig:intro}
\end{figure*}

The LegoNN decoder is the standard transformer decoder \cite{vaswani2017attention} with an added Ingestor (ING) component to consume input marginal distributions from preceding encoder modules. We propose two ingestor architectures; one that is differentiable, allowing for communicating gradients between modules, called the Weighted Embedding Ingestor (WEmb) and another discrete one that communicates a ranked list of hypotheses while keeping the modules gradient-isolated, called the Beam Convolution Ingestor (BeamConv). These ingestor components can be added on top of any module that accepts a distribution, so in a multi-module system (Eq. \ref{eq_multi_module}) these can be part of a LegoNN encoder. 

\paragraph{Weighted Embedding Ingestor (WEmb)}
The weighted embedding ingestor (WEmb) computes the expected embedding vector ($\mathbf{h}_{1:K}$) of the encoder distributions ($\mathbb{P}^{\mathcal{V}_{\text{CTC}}}_{1:K}$) per output time-step. Since these embedding vectors are formed out of local normalized conditionally independent CTC distributions, we need to re-encode the positional information in these embeddings. We combine the expected embedding vector ($\mathbf{h}_{1:K}$) with sinusoidal positional embedding (PE) before applying a few layers of self-attention transformer encoder blocks \cite{vaswani2017attention} to aggregate information across time-steps,
\begin{align}
    \mathbf{h}_{1:K} = \mathbb{P}^{\mathcal{V}^{(i-1)}_{\text{CTC}}}_{1:K} * W_{\text{Emb}};\,\,\,\, \mathbf{h}_{1:K} = \mathbf{h}_{1:K} + \operatorname{PE}(\mathbf{h}_{1:K}), \\
    \mathbf{h}_{1:K} = \operatorname{TransformerEncoder}(\mathbf{h}_{1:K}), \label{wemb_tsf}
\end{align}
where $\mathbb{P}^{\mathcal{V}^{(i-1)}_{\text{CTC}}}_{1:K}$ is the distribution over the vocabulary $\mathcal{V}^{(i-1)}_{\text{CTC}}$ of the previous module $i-1$, $W_{\text{Emb}}\in \mathbb{R}^{|\mathcal{V}_{\text{CTC}}^{(i-1)}|\times d}$ and $d$ is the input dimension for Eq. (\ref{wemb_tsf}) of the current module $i$.

The first operation, to compute the expected embedding, is equivalent to a 1-D convolution operation with a receptive field $\text{RF}$=1. When extended to larger receptive fields, WEmb offers the opportunity to learn local confusion patterns of the previous module: $\mathbf{h} = \text{Conv1D}(\mathbb{P}_{1:K}^{\mathcal{V}^{i-1}_{\text{CTC}}})$; with $\text{RF}$ $\geq$ 1.

\paragraph{Beam Convolution Ingestor (BeamConv)}
Rather than using the full output probability values $\mathbb{P}^{\mathcal{V}^{(i-1)}_{\text{CTC}}}_{1:K}$, the beam convolution ingestor (BeamConv) uses only the token indices of the $\operatorname{top-p}$ hypotheses for each position $k$ from the output of the preceding module. This creates an information bottleneck \cite{bottleneck99} in the model where gradients cannot be communicated,
\begin{align}
    \operatorname{top-p}(\mathbb{P}_{k}^{\mathcal{V}_{\text{CTC}}^{(i-1)}}) = \underset{A \subset \mathcal{V}_{\text{CTC}}^{(i-1)}, |A|=p}{\operatorname{argmax}} \sum_{a\in A} \mathbb{P}_{k}^{\mathcal{V}_{\text{CTC}}^{(i-1)}}(z=a), 
\end{align}
where $\mathbb{P}_{k}^{\mathcal{V}_{\text{CTC}}^{(i-1)}}(z = a)$ is probability of predicting the output token $a$ at k-th position for module $i-1$. The $\operatorname{top-p}$ indices are embedded into $d$ dimensional table, and, similar to WEmb ingestor, we apply positional embedding and self-attention transformer encoder blocks. Further, we can also use a 1-D convolution to aggregate local information,
\begin{align}
    \mathbf{r}_{1:K} &= \text{Embedding} \left( \operatorname{top-p}(\mathbb{P}^{\mathcal{V}^{i-1}_{\text{CTC}}}_{1:K}) \right), \\ 
    \mathbf{h}_{1:K} &= \text{Conv1D}(\mathbf{r}_{1:K}),\\
    \mathbf{h}_{1:K} &= \mathbf{h}_{1:K} + \text{PE}(\mathbf{h}_{1:K}),\\ \mathbf{h}_{1:K} &= \operatorname{TransformerEncoder}(\mathbf{h}_{1:K}),
\end{align}
where $\mathbf{r} \in \mathbb{R}^{T\times p \times d}$ when beam size=$p$ and RF $\geq$ 1 for input of $T$ time steps.



\section{Modularity Tests and Benefits of Modularity}
\label{sec:description_mod_tests_benefits}
In order to present the efficacy of LegoNNs, we subjected the LegoNNs to a variety of experiments that we describe in this section. These tests are designed to show that LegoNNs don't compromise on performance, are modular, and are flexible across tasks. Their results are shown in \Sref{sec:results}.

\subsection{Performance Tests}
\label{sec:perf_tests_sec4}
The benefits of modularity should not come at a cost to performance in individual tasks. To show that LegoNNs achieve competitive results, we compare their performance to the baseline monolithic encoder-decoder across benchmarks for speech recognition (ASR) and machine translation (MT).
\subsection{Modularity Tests}
\label{sec:modularity_tests_sec4}
To show that LegoNNs are indeed modular, we subject LegoNNs and the baseline monolithic encoder-decoder to various stress tests that check the modularity of these models. Similar to \cite{naik-etal-2018-stress}, these tests are designed to check if a system is modular or not. We consider three basic tests: 
\begin{enumerate}
    \item \emph{Random-Seed Swap} - Encoder or decoder modules trained from a different random seed should have the same functionality and hence be swappable.
    \item \emph{Architecture Swap} - Encoder or decoder modules trained with different architectures (including changes in size and number of parameters) should have the same functionality and hence be swappable.
    \item \emph{Modular Plug} - Modules trained in isolation should be able to plug-in to modules from previously trained LegoNN models that have the same interface. In this paper, we limit our scope towards training encoders in isolation and leave training modular decoders in isolation for future work.
\end{enumerate}
\subsection{Benefits of Modularity}
\label{sec:benefits_of_modularity_tests_sec4}
In order to present the benefits of modularity, we consider a practical scenario where practitioners need to build multiple ASR and MT tasks, as shown in \Fref{fig:intro}. In a situation where practitioners have a high resourced De-En machine translation model in their inventory, we show with LegoNNs they can save compute resources and leverage the well trained De-En decoder by applying the \emph{Modular Plug} to build systems for other MT and ASR tasks. In particular, we consider three scenarios:
\begin{enumerate}
    \item \emph{Romanian-English (Ro-En) MT} - In order to build a machine translation system for an under-resourced task like Romanian-English MT, practitioners using the LegoNN framework only need to build the Ro-En LegoNN encoder and re-use the De-En decoder. Additionally, they can also benefit the under-resourced Ro-En MT task, as the De-En decoder is trained on a larger corpus. 
    \item \emph{English ASR} - Practitioners can also use LegoNN decoders from an MT task for a different task on a different input modality. The OLC ensures that the expected encoder length for both MT and ASR task match, thereby making the De-En decoder re-usable.
    \item \emph{Different Domain English ASR} - We show that LegoNN modularity points are not limited to between encoders and decoders. While building an English ASR system on a different domain, practitioners can re-use components from both the English ASR and De-En MT systems.
\end{enumerate}
All the composed LegoNN models are end-to-end differentiable allowing further fine-tuning to improve performance if the practitioner has additional compute resources available.
\section{Experimental Setup}
\label{sec:exp_setup}
All our encoder-decoder models use the transformer architecture \cite{vaswani2017attention} implemented in the fairseq library \cite{ott2019fairseq} and run on DGX-1 nodes with 8 NVIDIA V100 GPUs. For both tasks, we apply LayerNorm \cite{ba2016layer} before every residual connection and a final one at the end of all transformer blocks. We use Adam optimizer~\cite{kingma2014adam} with $\text{eps}=1e^{-9}, \text{betas}=(0.9, 0.999)$, label smoothing=$0.1$, and a gradient clip norm $=5.0$. The model hyperparameters are detailed in Appendix \ref{sec:hyperparameters}.

\subsection{Speech recognition task}
\noindent\emph{Data:}\quad
For our speech recognition experiments, we follow the standard $300$ hours Switchboard (LDC97S62~\cite{godfrey1993switchboard}) setup, and the Switchboard (SWB) and CallHome (CH) subsets of HUB5 Eval2000 set (LDC2002S09~\cite{ldceval2kspeech}, LDC2002T43~\cite{ldceval2kctm}) for testing, which we use for the \emph{Performance Tests} (\ref{sec:perf_tests_sec4}). We follow the data preparation setup provided in ESPnet~\cite{espnet}, where we use $100$ and $2000$ target SentencePiece \cite{sentencepiece} units trained on the 300h text. We follow the same recipe for processing the TED-LIUM \cite{rousseauenhancing} and Europarl \cite{iranzo2020europarl} data, with phonemes generated using \cite{g2pE2019}, detailed in \Sref{sec:cross-task-data-setup}, which we use to present \emph{Benefits of Modularity} (\ref{sec:benefits_of_modularity_tests_sec4}).
We use the last model for inference with a beam size of $20$ and a length penalty of $1.0$. We do not use an external LM or joint decoding over the encoder and decoder outputs \cite{espnet}.\\[0.05in]
\noindent\emph{Model architecture:}\quad
Input features are processed using two 2-D convolution blocks with $3\times3$ kernels, $64$ and $128$ feature maps, respectively, $2\times2$ maxpooling and ReLU non-linearity. The baseline model uses transformers \cite{vaswani2017attention} with $16$ encoder blocks and $6$ decoder blocks, each with $1024$ dimensions, $16$ heads, $4096$ feed-forward units, and sinusoidal positional embeddings are added to the output of the convolutional context layers \cite{ConvTransformer}. 
For the LegoNN encoder-only model trained using the CTC loss, we use the same architecture as the encoder of the baseline model along with a length control unit which reduces the length of the input by a factor of $1.5$ with a maximum allowable length of $230$ time-steps. These positional embeddings are then passed through $6$ layers of transformer layers with cross-attention as described in \Sref{sec:olc}. The LegoNN decoder uses the same architecture as the baseline decoder. All Ingestor components (\Sref{sec:ing}) use RF=$1$, $3$ layers of transformers with $1024$ dimensions, $16$ heads and $4096$ feed-forward layer. The BeamConv ingestor uses K=$10$, and embedding size= $100$.\\[0.05in]
\noindent\emph{Training:}\quad
We use an average batch-size=$300$ utterances, weight decay=$1e^{-6}$, and lr=$1e^{-3}$ with $35k$ warm-up steps then exponentially decay to $5e^{-6}$ over $44k$ steps. We follow the SS policy of SpecAugment~\cite{specaugment} without time-warping. 

\subsection{Machine translation task}
\label{sec:machine-translation-task}
\noindent\emph{Data:}\quad
For the \emph{Performance Tests} (\ref{sec:perf_tests_sec4}), we train our models on the standard $4.5$M dataset from WMT En-De task, as used by \cite{vaswani2017attention, ott2018scaling}. 
We filter the training data to have a length ratio of $1.5$ with $250$ as max tokens. We test them on the \emph{newstest2011-2016} sets excluding the \emph{newstest2013} for the validation set. We use the shared $32$K BPE vocabulary \cite{BPE} provided by \cite{vaswani2017attention}. We average a moving window of $10$ checkpoints and pick the one with the best validation BLEU score. We use a beam size of $5$ and a length penalty of $0.6$ for decoding. The models are evaluated on case-sensitive tokenized BLEU with compound-splitting using \texttt{multi-bleu.pl}~\cite{multibleu}. For the \emph{Benefits of Modularity} experiments (\ref{sec:benefits_of_modularity_tests_sec4}), we train De-En and Ro-En models on WMT19 and WMT16 datasets, respectively. The data preparation for De-En MT models being used to compose with Ro-En MT encoder are detailed in \Sref{sec:cross-lingual-data-setup} and one being used to compose with ASR encoder is detailed in \Sref{sec:cross-task-data-setup}.\\[0.05in]
\noindent\emph{Model architecture:}\quad
The baseline model uses transformers \cite{vaswani2017attention} with $12$ encoder blocks and $6$ decoder blocks, each with $1024$ dimensions, $16$ heads, $4096$ feed-forward units, and sinusoidal positional embeddings. All embedding tables are shared across the model. To control for the extra parameters in the proposed LegoNN models, we added $6$ additional encoder blocks that improved the baseline model. Other strategies for using these parameters in the baseline model yielded inferior performance. 
The LegoNN encoder model uses $12$ transformer blocks with $1024$ dimensions, $8$ heads, and $2048$ feed-forward units, with OLC, upsampling the input length by a factor of $2$, applied to the second half of the encoder. The LegoNN decoder uses the same architecture as the baseline decoder. All Ingestor components (\Sref{sec:ing}) use RF $1$, $3$ layers of transformers with $1024$ dimensions, $16$ heads and $4096$ feed-forward units. The BeamConv ingestor uses K=$200$ and embedding size=$300$. Input embedding tables are shared with encoder, ingestor, and decoder tables where applicable.
\\[0.05in]
\noindent\emph{Training:}\quad
We use an average batch-size of $4000$ sentences, weight decay=$0.1$, and lr=$1e^{-3}$ with $35k$ warm-up steps followed by inverse square root decay for $45k$ steps. 

\subsection{Experimental Setup for Benefits of Modularity}

\subsubsection{Ro-En MT Modular Plug setup} 
\label{sec:cross-lingual-data-setup}

For our Ro-En transfer experiment, we used the WMT16 Ro-En and WMT19 De-En datasets. For WMT16 Ro-En, we used the data prepared by \cite{lee2018deterministic} which is already tokenized and lowercased.
For De-En, we excluded ParaCrawl from the standard WMT19 raw training set to obtain $7.4$M parallel translation pairs. We applied the same processing as for the Ro-En dataset by lowercasing and tokenizing the De-En set.
We prepared a single joint dictionary of $41000$ BPE units by combining the training text from both datasets. We filter the training data to have a length ratio of $1.5$ and $80$ as the max token length \cite{lee2018deterministic}. 
Following \cite{lee2018deterministic, marjan_axe}, we report tokenized BLEU scores on this dataset.
\\[0.05in]
\noindent\emph{Training:}\quad
As the Ro-En dataset contains only $610$K pairs, to achieve the best baseline performance, we reduce the size of the baseline model to avoid overfitting \cite{marjan_axe}. We use $6$ encoder and decoder layers with $512$ dimensions, $8$ heads, and $2048$ feed-forward units. We modify the Ro-En encoder-only LegoNN model accordingly by using transformer blocks with $512$ dimensions, $16$ heads, and $4096$ feed-forward units. 
For De-En LegoNN models along with the architecture described in \Sref{sec:machine-translation-task} we also experimented with larger models where the transformer encoder block has $16$ heads and $4096$ feed-forward units. We found that the larger BeamConv model performs better for this cross-lingual modular experiments.
\\[0.05in]
\noindent\emph{Fine-tuning:}\quad To finetune the composed LegoNN Ro-En WMT model we modify the learning rate to lr$=5e^{-6}$ and warm-up steps to $15k$. We run the training for around $4k$ steps to obtain the best validation perplexity.

\subsubsection{English ASR Modular Plug setups} 
\label{sec:cross-task-data-setup}
To demonstrate the third LegoNN scenario in \Fref{fig:intro}, we used the Europarl speech data to train our ASR task. 
We followed the data preparation for the ASR models described in \cite{iranzo2020europarl} by lowercasing, tokenizing, and stripping the punctuations from the text. We followed the same recipe when training models for the WMT19 De-En dataset, using the raw text from \Sref{sec:cross-lingual-data-setup}. To prepare the BPE target units, we trained a BPE model with  vocab size $2000$ on the Europarl train text and applied the prepared dictionary on the other sets of Europarl speech and De-En WMT data. We had an OOV rate of $0.096\%$ in the De-En WMT train data. 

For demonstrating the fourth LegoNN scenario in \Fref{fig:intro}, we used the TED-LIUM dataset \cite{rousseauenhancing} for training the pronunciation model module.
We processed the transcripts in the same way by lowercasing, tokenizing, and stripping the punctuation from the text data. We trained the $2000$ BPE target units on Europarl and TED-LIUM train text and applied the prepared dictionary to the other sets of Europarl, TED and De-En WMT. We had an OOV rate of $0.090\%$ in the De-En WMT train data. 

We use the grapheme-to-phoneme library described in \cite{g2pE2019} to get the target phonemes for the speech data used in both experiments. 
We report the final WER on the lowercased set to follow the standard ASR data setups.
\\[0.05in]
\noindent\emph{Training:}\quad
As the Europarl data contains only 70 hours, we reduced the size of the baseline model to avoid overfitting. We use transformers \cite{vaswani2017attention} with $12$ encoder and $6$ decoder blocks, each with $512$ dimensions, $8$ heads, $2048$ feed-forward units. 

For the LegoNN phoneme+pronunciation encoder-only model trained using CTC loss, we use a length control unit (for phonemes) which reduces the length of the input by a factor of $1.2$ with a maximum allowable length of $365$ time-steps. It outputs a marginal distribution over the phonemes, which is then passed to an Ingestor component (Wemb RF=$5$) followed by $6$ encoder layers and another output length control unit (for BPE output tokens) reducing the speech input by a factor $3.5$ with a maximum allowable length of $130$ time-steps. For the De-En MT model, we use the same architecture as \Sref{sec:cross-lingual-data-setup} but without sharing a dictionary between source and target. 
\\[0.05in]
\noindent\emph{Fine-tuning:}\quad To finetune the composed LegoNN Europarl ASR model we modify the peak learning rate to lr$=1e^{-4}$ and warm-up to a $1k$ steps. We run the training for $20k$ steps to get the best validation perplexity.

\begin{figure*}[t]
  \centering
    \includegraphics[width=\linewidth]{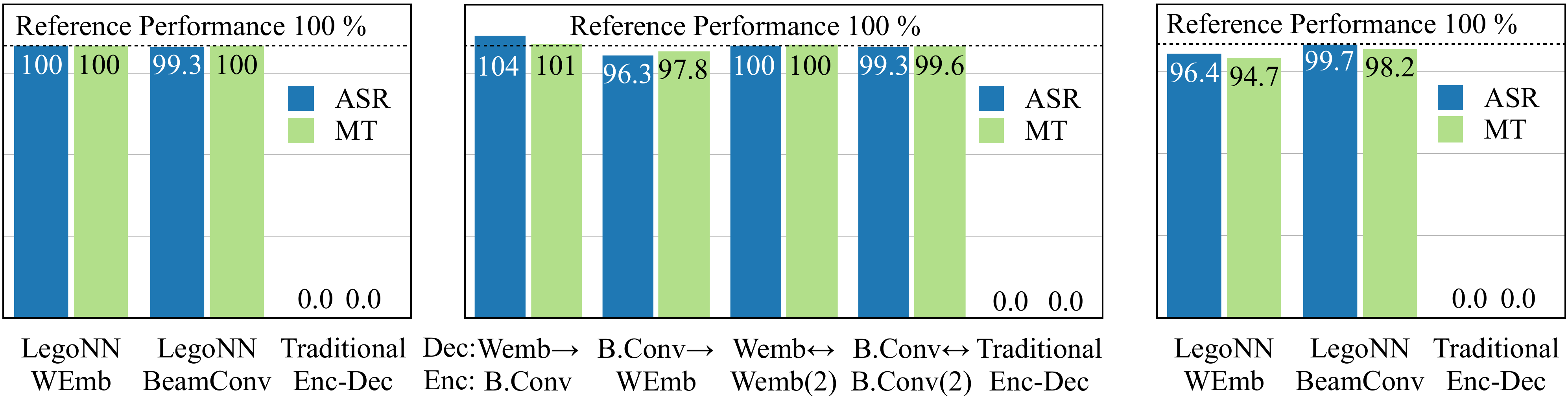}
    \caption{With reference performance of LegoNN models of 
    Tables~\ref{tab:lego_mt} and \ref{tab:lego_asr} (normalization to 100\% enables us to plot ASR and MT performances in the same graph), we show the interchangeability of LegoNN models compared to traditional Enc-Dec models under three conditions: swapping encoder and decoder modules of two models trained with different random seeds (left), swapping modules between LegoNN models with different architectures and trained with different ingestor types (middle), and matching arbitrary decoder modules with LegoNN encoder modules trained in-isolation (right).}
    \label{fig:resilience}
\end{figure*}

\section{Results}
\label{sec:results}
In this section we subject the LegoNN systems to the \emph{Peformance} and \emph{Modularity} tests introduced in \Sref{sec:perf_tests_sec4} and \Sref{sec:modularity_tests_sec4}. We also demonstrate the benefits of modularity towards reusing LegoNN modules between ASR and MT tasks, as presented in the four scenarios \Sref{sec:benefits_of_modularity_tests_sec4} and \Fref{fig:intro}.

\subsection{Performance of LegoNN models}
\label{sec:exp_performance}
First, we show the performance of LegoNN models on their original tasks, without sharing any modules. Tables \ref{tab:lego_mt} and \ref{tab:lego_asr} show the performance of the models trained \footnote{Table \ref{tab:lego_mt} and \ref{tab:lego_asr} report average scores on three random seeds} with the LegoNN procedure. 
For the WMT task, our best LegoNN model with the WEmb ingestor is only 0.8 BLEU behind \footnote{We believe this gap in LegoNN MT is due to the relatively lower quality of encoder-only MT models (compared to LegoNN encoder performance in ASR). With the recent advances in CTC-based MT encoder models, like using sequence level knowledge distillation \cite{imputer_MT,gu2021fully} this performance gap could be further reduced.} our strong baseline encoder-decoder model (better than the publicly available reference model by \cite{ott2018scaling}\footnote{\label{scaling_footnote}We downloaded the public model by \cite{ott2018scaling} to score the \emph{newstest2011-2016} test sets which weren't reported in their original paper.}) while being composed of modular reusable pieces (as we show in \Sref{sec:resilience} and \Sref{sec:exp_crosstasks}).
For the ASR task, our LegoNN models reach the same level of performance as the baseline encoder-decoder model. The good ASR performance of the gradient-isolated case of the BeamConv ingestor shows that the ASR task is amenable to the decomposition of the linguistic unit recognition and language generation components. \footnote{Using CTC models with language models is common for ASR \cite{miao2015eesen,amodei2016deep} and there can be some confusion regarding the relation of LegoNNs with them. We discuss this in \Sref{sec:text_only_lm}.}
\begin{table}[t]
      \caption{BLEU Scores ($\uparrow$) on WMT for LegoNN and baseline enc-dec MT models.}
        \label{tab:lego_mt}
    \centering
    \begin{tabular}{lcccc}
\toprule
\multirowcell{2}{MT Task} & \multicolumn{2}{c}{Loss Criterion} & \multicolumn{2}{c}{WMT En$\rightarrow$De}\\
      & CTC & CE & dev($\uparrow$) & test($\uparrow$) \\
\toprule
Scaling NMT \cite{ott2018scaling} & \xmark & \cmark & 27.3 & 28.0\footref{scaling_footnote} \\
\toprule
\multicolumn{5}{c}{Baseline Models (Our Implementation)}\\
\toprule
Baseline Enc-Dec & \xmark & \cmark & 27.6 & 28.3 \\
\toprule
\multicolumn{5}{c}{\textbf{LegoNN Models}} \\
\toprule
Encoder Only & \cmark & \xmark & 19.1 & 18.5 \\
Encoder + BeamConv Decoder   & \cmark & \cmark & 26.7 & 26.9 \\
Encoder + WEmb Decoder       & \cmark & \cmark & 27.2 & 27.5 \\
\bottomrule
\end{tabular}
\end{table}
\begin{table}[t]
      \caption{\% WER ($\downarrow$) on SWBD 300h (no LM) for LegoNN and baseline enc-dec ASR models.}
        \label{tab:lego_asr}
    \centering
    \begin{tabular}{lcccc}
\toprule
\multirowcell{2}{ ASR Task } & \multicolumn{2}{c}{Loss Criterion} & \multicolumn{2}{c}{Eval 2000 }  \\
      & CTC & CE & SWB($\downarrow$) & CH($\downarrow$)  \\
\toprule
LAS + SpecAugment \cite{specaugment} & \xmark & \cmark &  \hphantom{0}7.3\% & 14.4\% \\
IBM SWBD 300h \cite{tuske2020single} & \xmark & \cmark &  \hphantom{0}7.6\% & 14.6\% \\
ESPNET \cite{Karita2019ACS}  & \cmark & \cmark &  \hphantom{0}9.0\% & 18.1\% \\
Kaldi Hybrid system \cite{povey2016purely} & \multicolumn{2}{c}{LF-MMI} &  \hphantom{0}8.8\% & 18.1\% \\
\toprule
\multicolumn{5}{c}{Baseline Models (Our Implementation)} \\
\toprule
Baseline Enc-Dec & \xmark & \cmark & \hphantom{0}8.5\% & 18.0\% \\
\toprule
\multicolumn{5}{c}{\textbf{LegoNN Models}} \\
\toprule
Encoder Only  & \cmark & \xmark & 11.5\% & 24.1\% \\
Encoder + BeamConv Decoder   & \cmark & \cmark &  \hphantom{0}8.5\% & 18.2\% \\
Encoder + WEmb Decoder        & \cmark & \cmark &  \hphantom{0} 8.4\% & 18.2\% \\
\bottomrule
\end{tabular} 
\end{table}

\subsection{Modularity of LegoNN models}
\label{sec:resilience}
\Fref{fig:resilience} shows that LegoNN is modular; the decoder modules can be reused with encoders from other models for the same task \footnote{All statistics computed in \Fref{fig:resilience} report average scores across all possible combinations of mixing three random seeds from each module.}(section \Sref{sec:exp_crosstasks} presents cross-task performance). The 100\% reference level refers to the respective LegoNN and encoder-decoder models performance from ~\Tref{tab:lego_mt} and~\ref{tab:lego_asr}. Normalizing scores with respect to their reference performance allows us to plot ASR and MT systems in the same figure.
For both ASR and MT tasks, without fine-tuning, performance barely changes when modules are swapped between two different models trained with different random seeds or completely different architectures, even though these two models have different learning dynamics due to the use of different ingestor types (\Fref{fig:resilience}, left and middle). The right side of~\Fref{fig:resilience} presents the case where a LegoNN encoder module is trained in isolation of any decoder module using a CTC loss, for either ASR or MT, then matched with an arbitrary decoder module at inference time. This is the most challenging condition, especially for the MT task. Relying entirely on the indices of $\operatorname{top-p}$ hypotheses, rather than their floating-point marginal probabilities, the BeamConv ingestor shows more robustness when reused within the same task compared to the WEmb ingestor. The traditional encoder-decoder models, which are built without reusability in mind, fail completely under all these conditions (shown on the right side of the three sub-figures).

\subsection{Reusability of LegoNN modules across tasks}
\label{sec:exp_crosstasks}

LegoNN modules can be reused across tasks with no fine-tuning and with almost no degradation in performance. \Tref{tab:ro} shows an improvement in MT performance by 1.0 BLEU point when a Ro-En LegoNN encoder module is composed with a LegoNN decoder module that is trained on the De-En data (the second LegoNN scenario in \Fref{fig:intro}). 
\Tref{tab:eu} takes this a step further by composing an MT trained decoder (De-En data) with an ASR trained encoder on the Europarl speech dataset (the third LegoNN scenario in \Fref{fig:intro}). The BeamConv ingestor depends on a fixed beam size $p$ that cannot be changed after initial training of the decoder module. This explains the inferior performance of the BeamConv as compared to the WEmb ingestor (which uses the full marginal distribution) when reused in a new task.
\begin{table}[t]
        \caption{BLEU scores ($\uparrow$) on Ro-En MT task using a LegoNN model composed of a decoder from a De-En MT model and an encoder-only module trained on Ro-En data.}
        \label{tab:ro}
      \centering
        \begin{tabular}{lrr}
\toprule
MT Task & Ro$\rightarrow$En BLEU ($\uparrow$) & GPU Hours ($\downarrow$)\\ 
\midrule
Baseline Ro-En Enc-Dec & 34.0 & 144 \\
\midrule
LegoNN Encoder only & 30.7 & 120\footref{ctc_hours} \\
\multicolumn{2}{l}{\textsc{\underline{+ De-En WMT LegoNN Modules}}}\\
BeamConv Decoder & 33.0 & 0\footref{gpu_footnote} \\
WEmb Decoder & \textbf{35.0} & 0\footref{gpu_footnote} \\
\bottomrule
\end{tabular}
\end{table}
\begin{table}[t]
          \caption{\% WER ($\downarrow$) on Europarl ASR task using a LegoNN model composed of a decoder from a De-En MT model and an encoder-only module trained on Europarl ASR data.}
          \label{tab:eu}
      \centering
          \begin{tabular}{lrr}
\toprule
ASR Task & Europarl \% WER ($\downarrow$)  & GPU Hours ($\downarrow$)\\
\midrule
Baseline Europarl Enc-Dec & 18.4\%  & 22 \\
\midrule
LegoNN Encoder only  & 19.5\%  & 13\footref{ctc_hours} \\
\multicolumn{2}{l}{\textsc{\underline{+ De-En WMT LegoNN Modules}}}\\
BeamConv Decoder& 22.8\% & 0\footref{gpu_footnote} \\
WEmb Decoder& \textbf{18.4\%} & 0\footref{gpu_footnote} \\
\bottomrule
\end{tabular}

\end{table}
\begin{table*}[t]
  \centering
    \caption{\% WER on the Europarl test with ASR encoder decomposed into two modules. Modules trained on the TED-LIUM dataset are combined with those trained on Europarl and WMT to bring the overall WER of the LegoNN ASR system just 0.6\% from the baseline with no fine-tuning.}
    \label{tab:multiple_modules}
    \begin{tabular}{lr}
\toprule
ASR Task & Europarl \% WER ($\downarrow$) \\
\midrule
Baseline Europarl Enc-Dec & 18.4\%  \\
\midrule
TED. Phoneme Recognizer + TED. Pronunciation Model  & 26.7\% \\
Europarl Phoneme Recognizer + TED. Pronunciation Model & 20.5\% \\
Europarl Phoneme Recognizer + TED. Pronunciation Model + De-En WMT Decoder & \textbf{19.0\%}\\
\bottomrule
\end{tabular}
\end{table*}
\begin{table*}[t]
  \centering
    \caption{BLEU~($\uparrow$) on Ro-En WMT and \%WER~($\downarrow$) on Europarl ASR task for the LegoNN models before and after end-to-end fine-tuning of the model composed of modules from different tasks in \Tref{tab:ro}, \Tref{tab:eu} and \Tref{tab:multiple_modules}.}
    \label{tab:fine_tuning}
    \begin{tabular}{lrrr}
\toprule
Composed LegoNN Model & No fine-tuning & With fine-tuning & Metric \\
\midrule
\Tref{tab:ro}: Ro-En WMT Encoder + De-En WMT Decoder & 35.0 & \textbf{35.5} & BLEU~($\uparrow$)\\
\Tref{tab:eu}: Europarl ASR Encoder + De-En WMT Decoder  & 18.4 & \textbf{16.1} & \%~WER~($\downarrow$) \\
\Tref{tab:multiple_modules}: Europarl Phoneme Recognizer + TED. Pronun. Model + De-En WMT Decoder & 19.0 & \textbf{14.8} & \%~WER~($\downarrow$) \\
\bottomrule
\end{tabular}
\end{table*}
To demonstrate the flexibility of building sequence-to-sequence models with LegoNN, \Tref{tab:multiple_modules} shows the fourth LegoNN scenario in \Fref{fig:intro}.
A pronunciation modeling module trained on the TED-LIUM dataset is used with a phoneme recognition module trained on the Europarl dataset. 
Then, a decoder trained on the De-En WMT task is added to the ASR model to bring the final WER, with no fine-tuning updates, just 0.6\% from the baseline encoder-decoder model.  
The TED-LIUM dataset is used in this experiment because it is closer in speaking style to Europarl. Both the public TED talks and the parliament speeches exhibit similarities in speaking style and are not spontaneous like the Switchboard data. However, there is a clear mismatch in the domain which is apparent in the 26.7\% WER of the initial TED-LIUM system when evaluated on the Europarl test set. 
Utilizing LegoNN to develop models for new tasks benefits from the data-efficiency of encoder-only modules (Additional experiments regarding the data-efficiency of encoder-only models are shown in \Sref{sec:lowresource_legoNN}) and overall shorter development time, e.g., 13 vs 22 GPU-hours on Europarl ASR and 120 vs 144 GPU-hours on Ro-En WMT.~\footnote{\label{ctc_hours}GPU hours for encoder training can be improved further by using CuDNN based CTC implementation~\cite{cudnnctc}}~\footnote{\label{gpu_footnote}Composing LegoNN decoders with the LegoNN encoders, is a simple plug and play and does not require any fine-tuning steps. So no additional GPU hours are used for the decoder.}

\subsection{Fine-tuning of LegoNN models}
\label{sec:fine_tune}

So far, we showed matching or better results for LegoNN models composed of pre-trained modules with no fine-tuning. Given that LegoNN decoders with the WEmb ingestor preserve full differentiability, such LegoNN models composed of pre-trained modules can be fine-tuned toward the target task. \Tref{tab:fine_tuning} shows that fine-tuning the De-En decoder with the Ro-En encoder achieves another 0.5 BLEU point, leading to an improvement of 1.5 points over the baseline model. 
Although the non-fine-tuned two-module LegoNN model is better than the three-module one for the Europarl English ASR task, fine-tuning LegoNN models composed of two pre-trained modules achieved 12.5\% WER reduction, compared to a 19.5\% reduction for the three-module one. Fine-tuning the three modules helped reduce the domain mismatch while preserving the benefits of the TED-LIUM data.

\section{Additional Discussions around LegoNNs}
\subsection{Interpretable Interface in LegoNN models}
\label{appendix_inferface}
Each LegoNN module has a task, a clear performance metric, and a defined interpretable interface that allows one to assess the quality of individual modules. This gives a sense of the contribution of each module towards the overall performance of the task and helps in better debugging of the end-to-end systems. 
In \Tref{tab:interpretable_interface}, we show the loss and performance of the encoder modules of the LegoNN models that were presented in the table \ref{tab:lego_mt} and \ref{tab:lego_asr}. The performance of the encoder module was calculated using greedy decoding of the encoder marginal distributions without using an external language model.

\begin{table}[ht]
\caption{BLEU Scores ($\uparrow$) and \% WER ($\downarrow$) at the output of encoder modules of the LegoNN systems presented in Table \ref{tab:lego_mt} and \ref{tab:lego_asr}.}
\label{tab:interpretable_interface}
\centering
\begin{tabular}{lrrr}
\toprule
ASR Task & CTC & \multicolumn{2}{c}{Enc. WER ($\downarrow$)}  \\
      & Loss & SWB  & CH \\
\midrule
WEmb LegoNN      & 0.60 & 11.6\% & 24.2\% \\
BeamConv LegoNN   & 0.65 & 11.4\% & 24.4\% \\
\midrule
MT Task & CTC & \multicolumn{2}{c}{Enc. BLEU ($\uparrow$)}  \\
& Loss & dev & test \\
\midrule
WEmb LegoNN       & 2.51 & 18.9  & 18.4 \\
BeamConv LegoNN   & 2.56 & 18.2 & 17.8 \\
\bottomrule
\end{tabular} 
\end{table}

\subsection{Importance of encoder grounding using CTC for reusability} 
\label{sec:exp_encoder_grounding}
Table \ref{tab:model_enc_grounding} shows the results of training WEmb LegoNN MT models without an OLC unit or CTC loss at the encoder output. The initial system works fine but its components fail completely when used with another independently trained module.
\begin{table}[ht]
\caption{Importance of CTC encoder grounding for modularity}
\label{tab:model_enc_grounding}
\centering
\begin{tabular}{lr}
\toprule
& En $\rightarrow$ De ($\uparrow$)\\
LegoNN Models & \emph{newstest14}\\
\midrule
WEmb Ingestor & 29.2 \\
w/ Enc. Swap & 28.7 \\
\midrule
No CTC Loss at encoder output & 29.2 \\
w/ Enc. Swap & 0.0 \\
\bottomrule
\end{tabular}
\end{table}
\begin{table}[ht]
\caption{Effect of \% WER ($\downarrow$) of transferring a fully trained ASR LegoNN decoder module on encoder-only module trained on different amounts of data.}
\label{tab:lowresource}
  \centering
 \begin{tabular}{lcccc}
 \toprule
 & \multicolumn{2}{c}{10\% Data} & \multicolumn{2}{c}{30\% Data} \\
 ASR Task & SWB($\downarrow$) & CH($\downarrow$) & SWB($\downarrow$) & CH($\downarrow$)\\
 \midrule
 Baseline Enc-Dec & 129.7\% & 136.6\% & 20.4\% &	37.1\% \\
 \midrule
 LegoNN Encoder Module & 70.5\% & 80.5\% & 25.4\% & 43.8\% \\
 \multicolumn{5}{l}{\textsc{\underline{+ SWBD 300h LegoNN Modules}}}\\
 BeamConv Decoder & \textbf{40.7\%} & \textbf{59.6\%} & \textbf{17.5\%} & \textbf{34.5\%} \\ 
 WEmb Decoder    & 42.2\% & 63.4\% & 18.5\% & 34.7\% \\
 \bottomrule
 \end{tabular}
\end{table}

\subsection{Re-using LegoNN modules in low-resource conditions}
\label{sec:lowresource_legoNN}
Table \ref{tab:lowresource} shows the improvements observed when using LegoNN decoder modules with encoders trained with 10\% and 30\% of the switchboard ASR training data. Applying a decoder trained on the full data provided more than 50\% and 10\% relative improvement on average compared to the baseline encoder-decoder models, and 30\% and 25\% compared to a LegoNN encoder module when trained on the 10\% and 30\% low resource conditions, respectively. Given that this experiment shows module transfer between models trained on the same dataset, encoders are trained without the output length control unit because their output length distribution matches that of the decoder input.

\subsection{Incorporating Text-only Resources in LegoNNs}
\label{sec:text_only_lm}


The authors of \cite{miao2015eesen,amodei2016deep} have shown that ASR encoders trained with CTC loss can use language models trained with text-only resources to decode CTC distributions into words. These models are modular, but they are not trained end-to-end and cannot be conditioned on the error patterns of the encoder. 

For comparison with the LegoNN encoder-decoder model, we decoded the CTC distributions of LegoNN encoders presented in \Tref{tab:lego_asr} with a language model trained on the same data. The LegoNN encoder with WFST decoding using a language model~\cite{miao2015eesen} achieves 9.1\% and 19.4\% WER on the SWB and CH test sets averaged across 3 seeds, which is inferior to the 8.4\% and 18.2\% WER achieved by the LegoNN encoder-decoder model (\Tref{tab:lego_asr}). We expect this gap to be greater when comparing the MT translation models, where the performance gap between LegoNN encoders and LegoNN encoder-decoder models is larger (\Tref{tab:lego_mt}). 

Additionally, we can still utilize external language models while decoding LegoNN encoder-decoder models via shallow-fusion during beam-search \cite{espnet} and other techniques such as back-translation \cite{sennrich2016improving}. For future directions, the ability to train decoder-only LegoNN modules can allow training decoder modules on text-only data. For example, simulating CTC-like distributions with text-only data can allow decoder modules to accept those distributions and train on text-only data.

The LegoNN approach also offers a novel way to use additional data from various different tasks and languages (\Tref{tab:ro} and \Tref{tab:eu}). For example, a LegoNN encoder-decoder model with multiple encoders processing different modalities can train a decoder module with data from multiple tasks spanning various modalities. 

\subsection{Insights towards training LegoNN Models}
\label{sec:insights_training}
In this section, we discuss insights that can help better inform designers on their choices for building LegoNN systems.
\paragraph{Choice of interface distributions}
Enforcing modularity is not limited to the use of distributions trained using the CTC loss. Any system that outputs distributions over a chosen categorical space can be chained with another component that accepts those distributions. As mentioned in \Sref{sec:vocab}, we recommend using distributions that do not suffer from output label bias \cite{awnilabelbias}, as this can lead to error propagation into subsequent modules. We also believe that selecting interface probability distributions that only condition on input is better for a clear division of functionality between encoders and decoders and allows the ingestors in the subsequent modules to recover from the input errors made by the previous module.
\paragraph{Choice of interface vocabulary size}
In principle, the choice of interface vocabulary size should not affect the modularity of LegoNN systems, however, in practice, this can be an important consideration for various reasons outside of modularity. Following the discussions in \Sref{sec:vocab_generalizability}, the performance on public benchmarks is dependent heavily on the evaluation dataset, for example, the ASR toolkit \cite{espnet} uses sub-word units of size 2000 for Switchboard and 16000 for Librispeech benchmark datasets. 
Similarly, while a larger vocabulary size leads to shorter sequences and makes sequence modeling easier by making the predictions more context dependent, this contextual dependency can hurt the conditional independence and make CTC predictions domain specific. 
Due to this trade-off, in our experiments, we found an interface vocabulary size of around 4k to work the best across ASR and MT tasks.  

We would also like to note that as the models for speech and language processing are getting larger, the size of the vocabulary is converging between the two modalities. For example, the Whisper speech model \cite{radford2022robust} uses the same vocabulary as the GPT-2 language model \cite{radford2019language}. Additionally, the vocabulary choice is no longer dataset-specific but rather trained on large amounts of text to help with generalizability in the wild \cite{radford2019language, chowdhery2022palm}. We believe that such advances in the field will reduce the burden on practitioners regarding the choice of interface vocabulary size.
\paragraph{Choice of interface vocabulary units}
The units for the vocabulary depend entirely on the designer, depending on where the modules are being re-used, for example, if a module is being re-used only for speech tasks, then using phonemes as interface vocabulary is okay, but the designer might prefer sub-word units if it requires being used for tasks that require longer contexts like machine translation.

The choice of interface vocabulary can also determine what input information will be preserved or suppressed. For example, choosing discrete orthographic units for speech signals can capture content, but suppress other acoustic information, such as speaker characteristics, prosody, and emotions. Latent discrete units learned by self-supervised learning models of speech, on the other hand, have been shown to capture this acoustic information along with the content \cite{hayashi2020discretalk,borsos2022audiolm}. However, these units will not be able to interface with text-only systems, highlighting the importance of the designer's choice in deciding the reusable components.

\paragraph{Choice of up-/down-sampling rate in OLC}
Output Length Controller (OLC) is an important component to be considered when modules are being used in dramatically different sequence tasks, for example, ASR and MT. When used in a single domain, that is if the expected input lengths of the decoder do not vary drastically then OLC is not required. OLC ensures that the CTC distributions are being learned over similar input-to-output alignment lengths for different tasks. 
\paragraph{Choice of Wemb and BeamConv LegoNN decoder}
As discussed in \Sref{sec:exp_crosstasks}, we believe that Wemb is a more robust choice when using LegoNN models in different domains and tasks, as it functions over the full marginal distribution to compute an expected embedding vector, while the top-p confusions in BeamConv can differ drastically across tasks and domains. The BeamConv, on the other hand, offers an interesting approach towards ingesting top-p confusions, which makes them useful for error analysis and provides capabilities to inject external knowledge into the confusion lattice.
\paragraph{Choice of language pair in LegoNN MT}
In the LegoNN setup discussed in \Sref{sec:benefits_of_modularity_tests_sec4}, we require the target language of the re-usable modules to be the same. There are no constraints on the source language and we do not expect the full LegoNN encoder-decoder performance to be affected by this compared to traditional monolithic models. Recent works have shown CTC-based modeling to work well for various different source languages \cite{yan2022ctc} and we would like to extend this work with LegoNNs to build multilingual any to English translation systems with re-usable English LegoNN decoders.

\section{Related work}
\label{sec:prior}
This work is related to the large body of work on probabilistic modeling for ASR and MT \cite{jelinek1997statistical, mohri2002weighted, brown1990statistical} where predictors produce normalized probabilities that can be easily combined. However, these probabilistic models only combine output scores while individually optimizing modules producing discrete sequences as opposed to chaining modules while preserving their full differentiability as in LegoNN models. Hierarchical mixture of experts \cite{jordan1994hierarchical} and graph transformer networks \cite{bottou1997global} motivated this work, however, the first does not ground intermediate representations and the second communicates them in the form of directed graphs, which can be computationally expensive.

The proposed LegoNN procedure builds on several research efforts for sequence-to-sequence learning. Encoder-decoder models for machine translation~\cite{oxford_s2s_2013, sutskever2014sequence, bahdanau2014neural, vaswani2017attention} and speech recognition~\cite{chan2016listen,bahdanau2016end} form the basis of this work. 
Although CTC loss~\cite{graves2006connectionist} was first applied to speech recognition~\cite{ctc_2013, sak2014long, deep_speech, imputer_ASR}, more recently it was also shown to be effective for machine translation~\cite{libovicky-helcl-2018-end, imputer_MT}. This encourages us to utilize the CTC loss for enforcing the intermediate vocabulary in LegoNN. 
Other non-autoregressive sequence to sequence mapping methods~\cite{Jiatao_SF, marjan_emnlp, marjan_axe, jhu_shinji} are potential alternatives. 
The CTC loss has been combined with the cross-entropy loss in encoder-decoder speech recognition systems to encourage monotonic alignment between input and output sequences~\cite{suyounkim, Karita2019ACS}. Different from LegoNN, their decoder attends over the encoder hidden output representations, maintaining their tight coupling.

There have been many proposals in the fields of vision, robotics and reasoning for inducing a modular structure on the space of learned concepts either through hierarchically gating information flow or via high-level concept blueprints~\cite{andreas2016neural, devin16, ranzato19} to enable zero- and few-shot transfer learning~\cite{andreas2017modular, socher2013zero, Gupta2020Neural, pathak2019learning}. 

\begin{table*}[ht]
    \caption{Individual year tokenized (tok.) and detokenized (detok.) BLEU Scores ($\uparrow$) on WMT for LegoNN and baseline enc-dec MT models.}
    \label{tab:individual_bleu_perf}
  \centering
    \begin{tabular}{lcccccccccccc}
\toprule
\multirowcell{2}{MT Task} & \multicolumn{12}{c}{WMT En$\rightarrow$De ($\uparrow$)}\\
      & \multicolumn{2}{c}{\emph{newstest11}} & \multicolumn{2}{c}{\emph{newstest12}} & \multicolumn{2}{c}{\emph{newstest13}} & \multicolumn{2}{c}{\emph{newstest14}} & \multicolumn{2}{c}{\emph{newstest15}} & \multicolumn{2}{c}{\emph{newstest16}} \\
      & tok. & detok. & tok. & detok. & tok. & detok. & tok. & detok. & tok. & detok. & tok. & detok. \\
\toprule
Scaling NMT \cite{ott2018scaling}~\footref{scaling_footnote_2} & 22.7 & 22.3 & 23.1 & 23.0 &	27.3 & 26.8 & 29.8 & 29.2 &	32.2 & 31.8 & 35.0 & 34.8 \\
\toprule
\multicolumn{13}{c}{Baseline Models (Our Implementation)}\\
\toprule
Baseline Enc-Dec & 23.3 & 22.8 & 23.3 & 23.1 &	27.6 & 27.2 & 29.8 & 29.0 & 31.9 & 31.5 & 35.6 & 35.1\\
\toprule
\multicolumn{13}{c}{\textbf{LegoNN Models}} \\
\toprule
Encoder Only & 16.0 & 15.8 & 15.7 & 15.7 &	19.1 & 19.0 & 18.0 & 17.8 &	20.3 & 20.2 & 21.6 & 21.7\\
Encoder + BeamConv Decoder & 22.4 & 21.9 & 22.4 & 22.3 &	26.7 & 26.3 & 27.3 & 26.6 & 29.8 & 29.5 & 33.4 & 33.0  \\
Encoder + WEmb Decoder & 22.7 & 22.3 & 22.8 & 22.6 & 27.2 & 26.8 & 28.3 & 27.6 & 30.3 & 30.0 & 34.3 & 33.9 \\
\bottomrule
\end{tabular}

\end{table*}

\section{Conclusion}
\label{sec:conc}
We presented the LegoNN procedure for constructing encoder-decoder models that are composed of reusable modules. LegoNN models perform competitively to the best encoder-decoder ASR and MT models on large-scale benchmarks. A key to reusable modules is a pre-defined vocabulary that is shared between many tasks across which modules can be reused. Without any fine-tuning steps, a LegoNN decoder trained for the De-En WMT task can replace an ASR decoder module without any impact on performance and provide better generation quality for a Ro-En WMT task. When fine-tuned for a few thousand steps, LegoNN models composed from multiple tasks and domains improve the Ro-En WMT baseline model by 1.5 BLEU points and provide up to 19.5\% WER reductions on the Europarl English ASR task. Our future directions will focus on combining the flexibility of LegoNN models with the impressive performance of encoder pre-training methods like BERT, and investigating zero-shot learning scenarios for speech translation which relies on a combination of ASR and MT modules. 

\section*{Broader Impact}
The software industry made great strides in building independent, reusable libraries that can be developed once and reused across a wide range of applications. This paper takes one step towards bringing sequence-to-sequence neural models closer to computer programs with reusable components. This line of research can provide a way for the research community and industry to build more complex neural systems while preventing an explosion in computational training costs. 


%

\appendices

\section{Detokenized BLEU performance of our LegoNN MT models}
\label{sec:sacrebleu}
\Tref{tab:sacrebleu_performance} shows the detokenized BLEU performance of the MT models used in \Tref{tab:lego_mt}.
\footnote{\label{scaling_footnote_2}We downloaded the public model of \cite{ott2018scaling} to score the \emph{newstest2011-2016} test sets which weren't reported in their original paper.} We used the detokenizer from mosesdecoder~\footnote{\url{https://github.com/moses-smt/mosesdecoder/blob/master/scripts/tokenizer/detokenizer.perl}} 
to detokenize the model output and SacreBLEU~\cite{post_bleu}
to calculate the BLEU score with the following hash -  \texttt{BLEU+case.mixed+lang.en-de+numrefs.1+\\smooth.exp+test.\{wmt11|wmt12|wmt13|wmt14/\\full|wmt15|wmt16\}+tok.13a+version.1.2.9}

\section{BLEU performance on individual years for our LegoNN MT models}
\Tref{tab:individual_bleu_perf} shows the BLEU performance (tokenized and detokenized BLEU) from individual years (\emph{newstest11-16}) for the MT models used in \Tref{tab:lego_mt}. 

\begin{table}[ht]
    \caption{SacreBLEU ($\uparrow$) on WMT for LegoNN and baseline enc-dec MT models.}
    \label{tab:sacrebleu_performance}
  \centering
    \begin{tabular}{lcccc}
\toprule
\multirowcell{2}{MT Task} & \multicolumn{2}{c}{Loss Criterion} & \multicolumn{2}{c}{WMT En$\rightarrow$De}\\
      & CTC & CE & dev($\uparrow$) & test($\uparrow$) \\
\toprule
Scaling NMT \cite{ott2018scaling} & \xmark & \cmark & 26.8 & 28.2~\footref{scaling_footnote_2} \\
\toprule
\multicolumn{5}{c}{Baseline Models (Our Implementation)}\\
\toprule
Baseline Enc-Dec & \xmark & \cmark & 27.2 & 28.3 \\
\toprule
\multicolumn{5}{c}{\textbf{LegoNN Models}} \\
\toprule
Encoder Only & \cmark & \xmark &  19.0 & 18.2  \\
Encoder + BeamConv Decoder   & \cmark & \cmark & 26.3 & 26.7 \\
Encoder + WEmb Decoder       & \cmark & \cmark & 26.8 & 27.3 \\
\bottomrule
\end{tabular}
\end{table}

\section{Training hyperparameters for ASR and MT task}
\label{sec:hyperparameters}
\Tref{tab:hp-pretrain-asr} and \Tref{tab:hp-pretrain-mt} contain the training parameter details for the LegoNN model.

\begin{table}[ht]
  \caption{Hyperparameters for training LegoNN model ASR task.}
  \label{tab:hp-pretrain-asr}
  \centering
  \begin{tabular}{lr}
    \toprule
    Hyperparameter & Value \\
    \midrule
    Hidden Dropout & $0.15$ \\
    Attention dropout & $0.15$ \\
    Activation dropout & $0.15$ \\
    Ingestor attention dropout & $0.15$ \\
    Batch size & $300$ utt.\\
    LR schedule & tristage \cite{specaugment} \\
    Start learning rate & $1e^{-6}$ \\
    Max learning rate & $1e^{-3}$ \\
    End learning rate & $5e^{-6}$ \\
    Number of steps & $80$K \\
    Warmup steps & $35$K \\
    Hold steps & $1$K \\
    Adam eps  & $1e^{-9}$ \\
    Adam betas  & ($0.9$, $0.999$)\\
    Weight decay & $1e^{-6}$ \\
    \bottomrule
  \end{tabular}
\end{table}

\begin{table}[ht]
  \caption{Hyperparameters for training LegoNN model MT task.}
  \label{tab:hp-pretrain-mt}
  \centering
  \begin{tabular}{lr}
    \toprule
    Hyperparameter & Value \\
    \midrule
    Hidden Dropout & $0.3$ \\
    Attention dropout & $0.3$ \\
    Activation dropout & $0.3$ \\
    Ingestor attention dropout & $0.1$ \\
    Batch size & $4$K sent.\\
    LR schedule & inv. sqrt. \cite{ott2018scaling} \\
    Start learning rate & $1e^{-6}$ \\
    Max learning rate & $1e^{-3}$ \\
    Number of steps & $150$K \\
    Warmup steps & $35$K \\
    Adam eps  & $1e^{-9}$ \\
    Adam betas  & ($0.9$, $0.999$)\\
    Weight decay & $0.1$ \\
    \bottomrule
  \end{tabular}
\end{table}



\section*{Acknowledgment}
The authors would like to thank Siddhant Arora, Brian Yan, Maria Ryskina, Shruti Rijhwani and Dan Berrebbi for their valuable feedback. We would also like to acknowledge and remember the late Sujeath Pareddy, who suggested the connection with Legos for our proposed framework.

\ifCLASSOPTIONcaptionsoff
  \newpage
\fi



%

\bibliographystyle{IEEEtran}
\bibliography{myplain,2021_modularity}




%








\end{document}